\documentclass[letterpaper, 10 pt, conference]{ieeeconf}
\IEEEoverridecommandlockouts
\usepackage{epsfig}
\usepackage{amssymb}
\usepackage{cite}
\usepackage{booktabs}
\usepackage{multirow}
\usepackage{mathtools}
\usepackage{bm}
\usepackage{algorithm}
\usepackage{comment}
\usepackage{color}
\usepackage{setspace}

\newcommand{\figlab}[1]{\label{fig:#1}}
\newcommand{\figref}[1]{Fig.~\ref{fig:#1}} % Figure

\newcommand{\forlab}[1]{\label{for:#1}}
\newcommand{\etal}{\textit{et al.~}}

\definecolor{green}{rgb}{0.01, 0.5, 0.01}

\usepackage{subcaption}
\usepackage{url}
\usepackage{threeparttable}

\usepackage{amsmath}
\usepackage[noend]{algorithmic}
\usepackage{array}

\begin{document}
\title{Active Vapor-Based Robotic Wiper}
\author{Takuya Kiyokawa$^{*}$, Hiroki Katayama, Jun Takamatsu, and Kensuke Harada
\thanks{T. Kiyokawa and K. Harada are with the Department of Systems Innovation, Graduate School of Engineering Science, Osaka University, Toyonaka, Osaka, Japan. $^{*}$Corresponding author: {\tt\small kiyokawa@sys.es.osaka-u.ac.jp}}%
\thanks{H. Katayama and J. Takamatsu are with the Division of Information Science, Nara Institute of Science and Technology (NAIST), Japan.}%
}

% make the title area
\maketitle

\vspace{-20pt}
\begin{abstract}
This paper presents a method for estimating normals of mirrors and transparent objects challenging for cameras to recognize. We propose spraying water vapor onto mirror or transparent surfaces to create a diffuse reflective surface. Using an ultrasonic humidifier on a robotic arm, we apply water vapor to the target object's surface, forming a cross-shaped misted area. This creates partially diffuse reflective surfaces, enabling the camera to detect the target object's surface. Adjusting the gripper-mounted camera viewpoint maximizes the extracted misted area's appearance in the image, allowing normal estimation of the target surface. Experiments show the method's effectiveness, with RMSEs of azimuth estimation for mirrors and transparent glass approximately $4.2^{\circ}$ and $5.8^{\circ}$, respectively. Our robot experiments demonstrated that our robotic wiper can perform contact-force-regulated wiping motions to clean a transparent window, akin to human performance.
\end{abstract}

\IEEEpeerreviewmaketitle

\section{Introduction}
For robots to be useful in home environments, they must recognize a wide range of objects. The robot must identify target object surfaces accurately to avoid damaging fragile items. Recognizing the plane normal of mirrors and transparent objects is challenging due to specular reflection or light transmission on non-Lambertian surfaces. To enable robots to clean windows and mirrors, a system that identifies and wipes these surfaces without specific machines~\cite{Miyake2006,Imaoka2010} is needed.

This study proposes using water vapor to enhance the visibility of mirrors and transparent surfaces for plane normal estimation. This is the first attempt to develop a robotic wiper system based on the \textit{Active Vapor} method, shown in~\figref{system}. The system creates a misted area by spraying water vapor with an ultrasonic humidifier attached to a robot arm gripper. The robot then wipes the surfaces based on the estimated plane normal with regulated contact force.
Experiments demonstrated that the robotic wiper accurately estimated plane normals for mirrors and transparent glass and performed effective wiping. The method estimated the azimuth angle with errors less than $4.2^{\circ}$ for mirrors and $5.8^{\circ}$ for transparent glass.
%%%%%%%%%%%%%%%%%%%%%%%%%%%%%%%%%%%%%%%%%%%%%%%%%%
\begin{figure}[t]
  \centering
  \begin{minipage}[t]{0.48\linewidth}
    \centering
    \includegraphics[width=\linewidth]{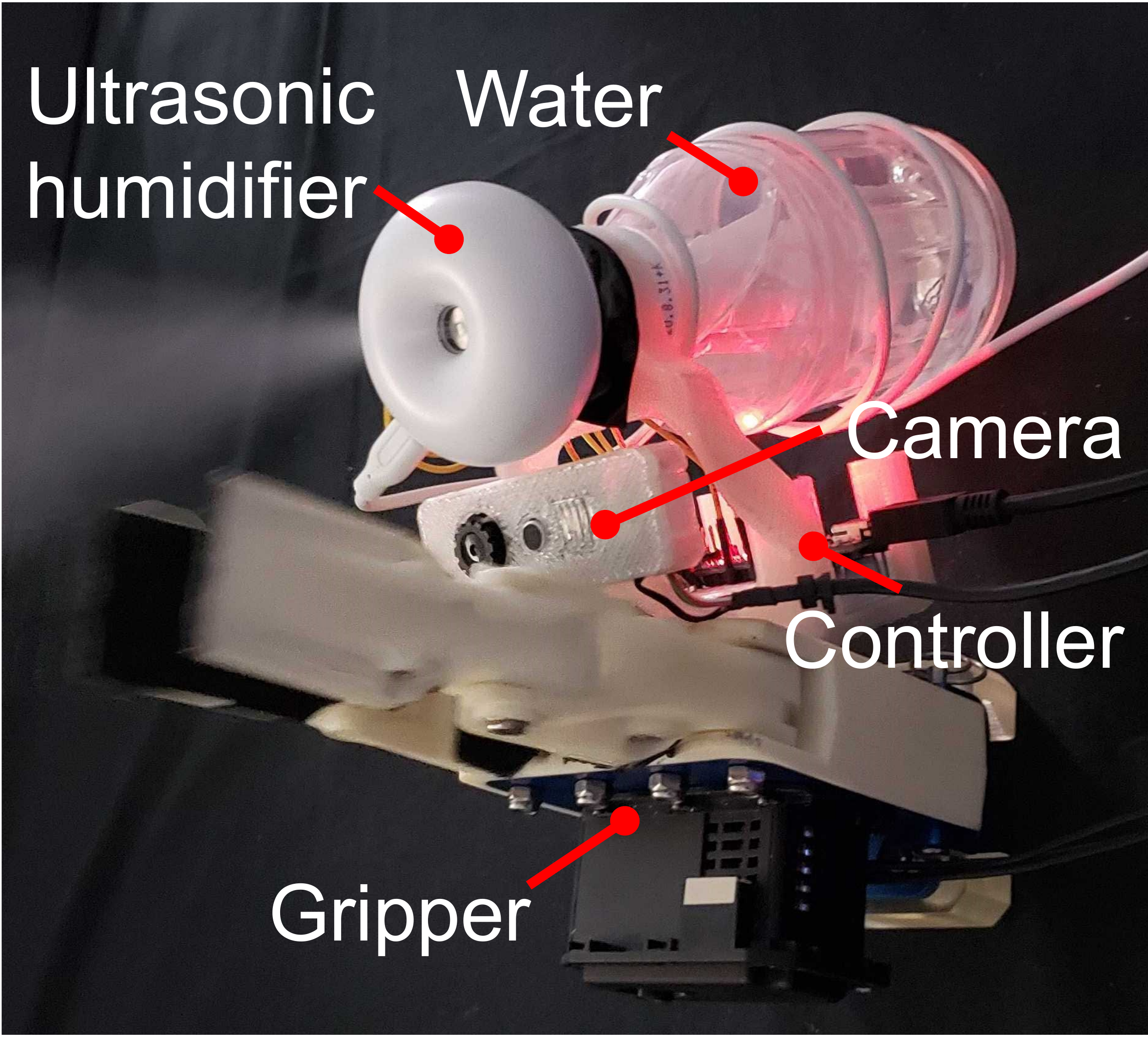}
    \subcaption{\small{\textit{Active Vapor} system}}
  \end{minipage}
  \begin{minipage}[t]{0.48\linewidth}
    \centering
    \includegraphics[width=\linewidth]{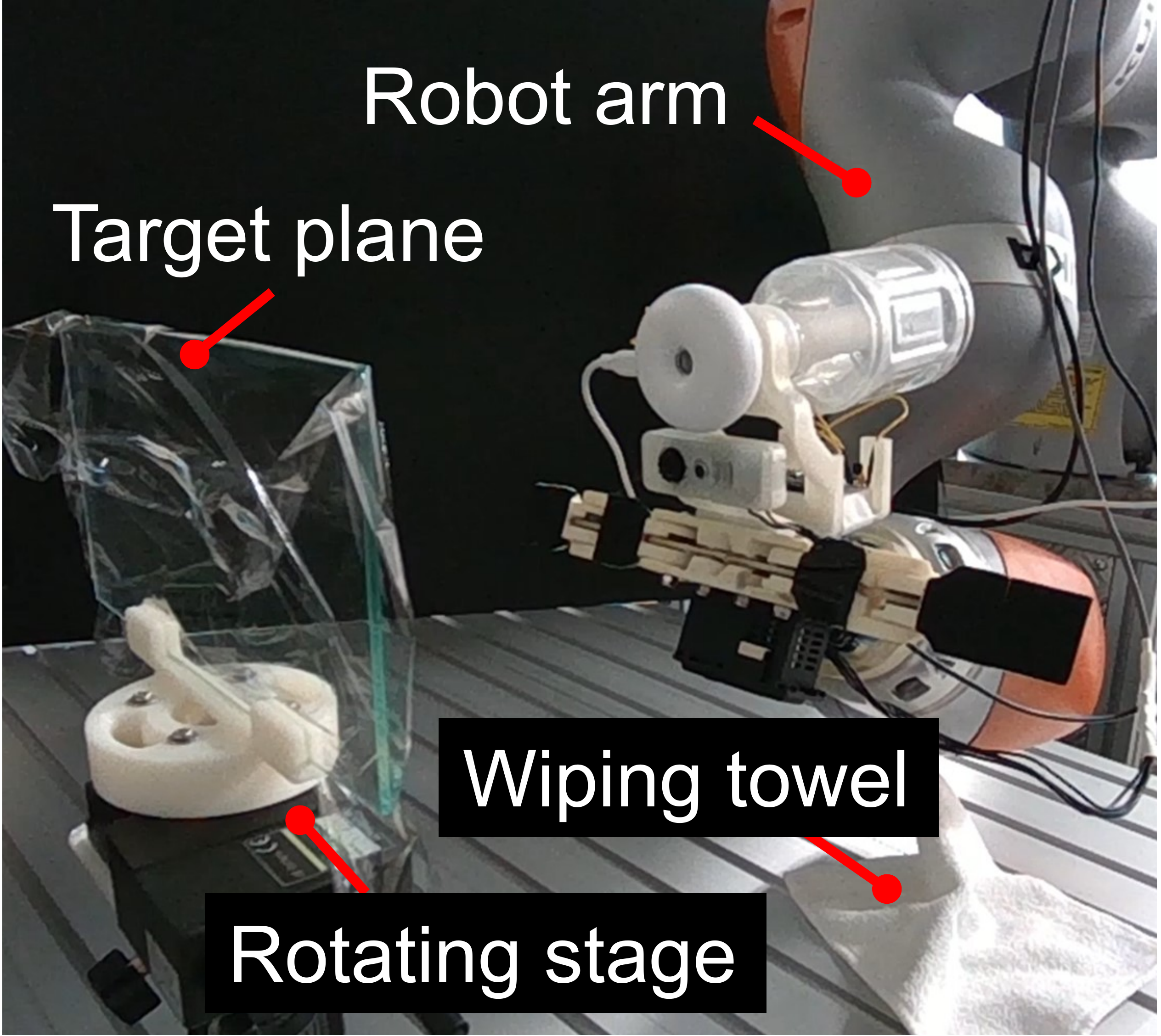}
    \subcaption{\small{Experimental wiper}}
  \end{minipage}
  \caption{\small{Robotic window-wiping system based on \textit{Active Vapor}. The system has three functions: spraying water vapor, grasping an object, and capturing an image.}}
  \figlab{system}
\end{figure}
%%%%%%%%%%%%%%%%%%%%%%%%%%%%%%%%%%%%%%%%%%%%%%%%%%

\section{Active Vapor-Based Normal Estimation}
%%%%%%%%%%%%%%%%%%%%%%%%%%%%%%%%%%%%%%%%%%%%%%%%%%%%%%%%%%%%
\begin{figure}[t]
    \centering
    \begin{minipage}[t]{0.58\linewidth}
      \centering
      \includegraphics[width=\linewidth]{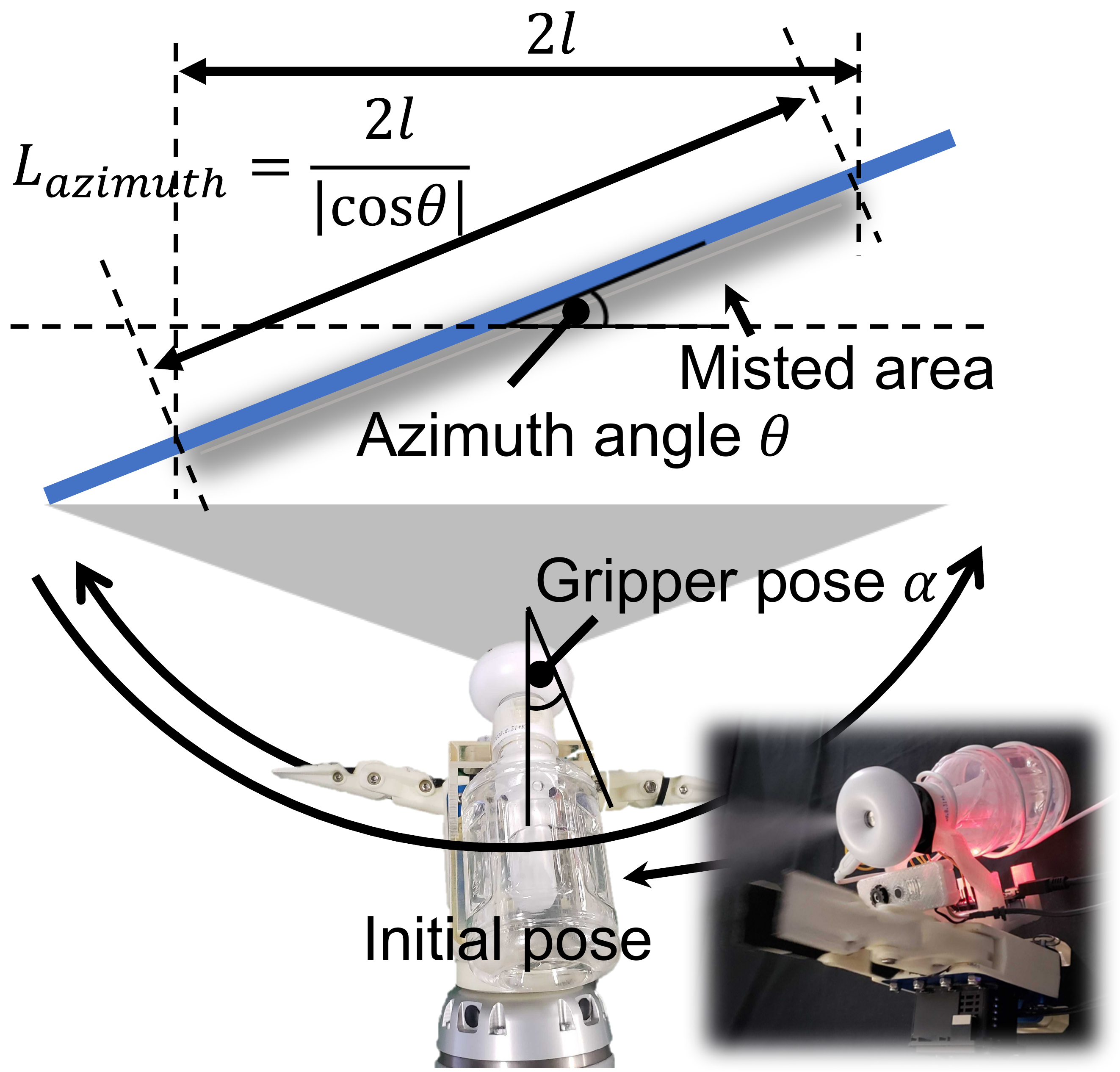}
      \subcaption{\small{Azimuth estimation}}
    \end{minipage}
    \begin{minipage}[t]{0.4\linewidth}
      \centering
      \includegraphics[width=\linewidth]{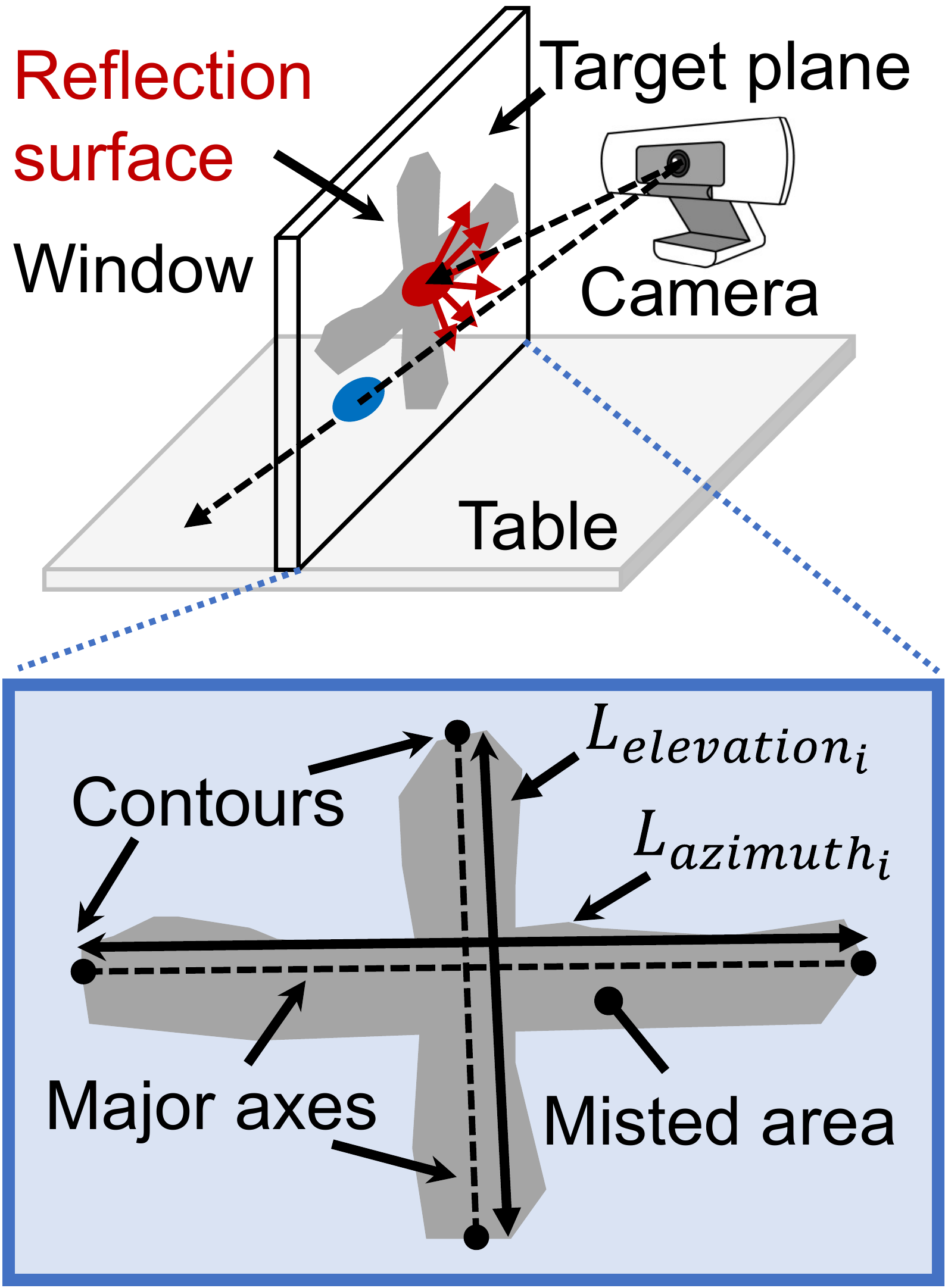}
      \subcaption{\small{Reflection surface}}
    \end{minipage}
    \caption{\small{Normal estimation method. (a) shows an overview of the gripper motion and geometries between the gripper and the misted plane. The top part of (b) shows the reflection surface that appears on the target plane. The bottom part of (b) shows the geometry of the cross-shaped misted area viewed from the viewpoint $i$.}}
    \figlab{method}
\end{figure}
%%%%%%%%%%%%%%%%%%%%%%%%%%%%%%%%%%%%%%%%%%%%%%%%%%

\subsection{Estimating Normal Direction from Misted Cross-Shape}
The misted area appears different from its surroundings due to its blurred, diffusely reflected nature. By shaping the misted area into a cross, we can estimate the normal direction from the line segment lengths, making precise boundaries unnecessary. A vapor-covered line segment suffices for normal estimation, suggesting traditional image segmentation methods are suitable.

The azimuth angle $\theta$ and elevation angle $\phi$ can be estimated based on the tilt angle geometry of the target plane and the cross-shaped misted area. As both angles are calculated similarly, the azimuth angle calculation is exemplified. \figref{method}~(a) illustrates the azimuth angle geometry. If the vertical and horizontal axis lengths in the cross-shaped misted area are $2l$, determined by the robot arm trajectory length, the angle value is calculated accordingly.
%%%%%%%%%%%%%%%%%%%%%%%%%%%%%%%%%%%%%%%%%%%%%%%%%%
\begin{equation}
    \theta = \pm{\cos^{-1}{\frac{2l}{L_{azimuth}}}}.
\end{equation}
%%%%%%%%%%%%%%%%%%%%%%%%%%%%%%%%%%%%%%%%%%%%%%%%%%
This solution is not uniquely determined. It is assumed that the calculated misted area length $L_{{azimuth}}$ is maximized when the target plane is perpendicular to the camera's image-capturing direction. The azimuth angle is estimated using this geometric constraint.

First, by adjusting the camera viewpoint with the robot arm after water vapor spraying, we identify the camera direction $i$, where $L_{{azimuth}_i}~(i=1,2,..,n)$ reaches its maximum, with $n$ representing the number of viewpoint adjustments made by the robot arm. Second, the angle between the estimated and initial directions before adjustment is determined as the estimated azimuth $\hat{\theta}$.
%%%%%%%%%%%%%%%%%%%%%%%%%%%%%%%%%%%%%%%%%%%%%%%%%%%%%%%%%%%%
\begin{figure}[t]
    \centering
    \includegraphics[width=.75\linewidth]{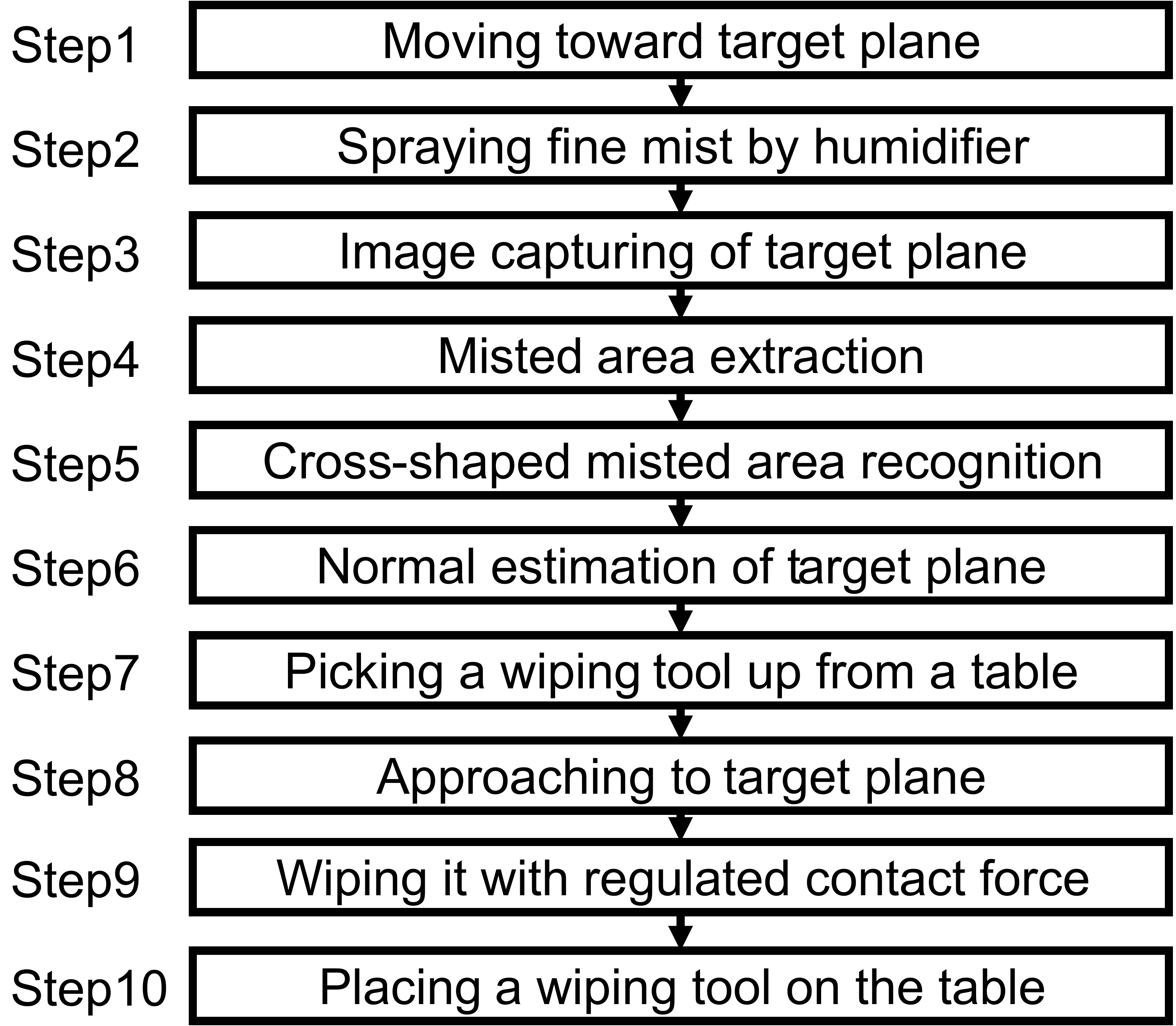}
    \caption{\small{Normal estimation of mirrors and transparent objects.}}
    \figlab{process}
\end{figure}
%%%%%%%%%%%%%%%%%%%%%%%%%%%%%%%%%%%%%%%%%%%%%%%%%%%%%%%%%%%%
As depicted in \figref{method}~(b) (top), $L_{{azimuth}_i}$ is derived by first finding the intersections of the major axes with the misted area's contours, then measuring the distance between these intersections.

With the estimated elevation angle $\hat{\phi}$, the normal $\bm{N}$ at the sprayed plane's initial position is calculated as
%%%%%%%%%%%%%%%%%%%%%%%%%%%%%%%%%%%%%%%%%%%%%%%%%%%%%%%%%%%%
\begin{equation}
    \bm{N} = (\cos{\hat{\phi}} \cos{\hat{\theta}},~\cos{\hat{\phi}} \sin{\hat{\theta}} ,~\sin{\hat{\phi}}).
\end{equation}
%%%%%%%%%%%%%%%%%%%%%%%%%%%%%%%%%%%%%%%%%%%%%%%%%%%%%%%%%%%%
\subsection{Spraying Water Vapor}
\figref{process} outlines the proposed method. Water vapor is sprayed as the robot gripper nears the target plane. The robot arm moves the vapor to create a cruciform misted area. This cross-shaped mist helps determine azimuth and elevation angles from the line lengths.

The robot arm executes reciprocating linear motions up/down and left/right from the initial position to form the misted cross. Post-spraying, the robot arm moves along arcs at a speed ensuring the arc movement is completed in less than $T_{completion}[{\rm s}]$, satisfying the following equation:
%%%%%%%%%%%%%%%%%%%%%%%%%%%%%%%%%%%%%%%%%%%%%%%%%%%%%%%%%%%%
\begin{equation} \forlab{tcomp}
    \frac{r\beta}{v_{wrist}} < T_{completion},
\end{equation}
%%%%%%%%%%%%%%%%%%%%%%%%%%%%%%%%%%%%%%%%%%%%%%%%%%%%%%%%%%%%
where $r$ [mm] is the distance to the target plane, $\beta$ [rad] is the rotation angle of the robot gripper when estimating the normal after completing water vapor spraying, and $v_{wrist}$ [mm/s] is the circumferential speed at that time.

\section{Wiping System Implementation}
\subsection{Active Vapor System}
%%%%%%%%%%%%%%%%%%%%%%%%%%%%%%%%%%%%%%%%%%%%%%%%%%%%%%%%%%%%
\begin{figure*}[t]
    \centering
    \begin{minipage}[t]{0.495\linewidth}
      \centering
      \includegraphics[width=\linewidth]{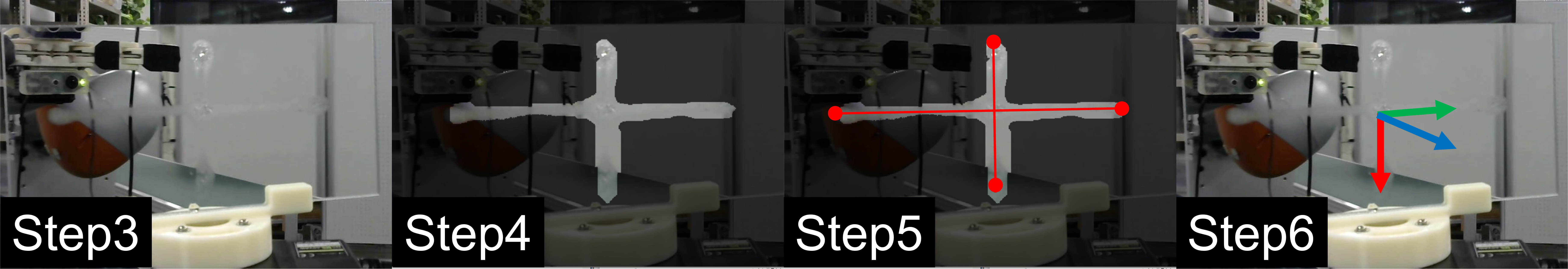}
      \subcaption{\small{Normal estimation process}}
    \end{minipage}
    \begin{minipage}[t]{0.495\linewidth}
      \centering
      \includegraphics[width=\linewidth]{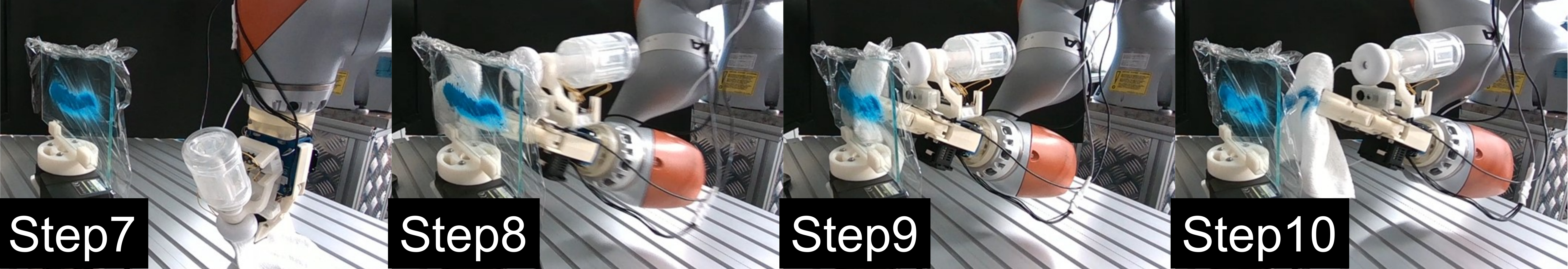}
      \subcaption{\small{Wiping motion}}
    \end{minipage}
    \caption{\small{ From normal estimation to wiping motion execution: (a) The system estimated the misted mirror surface. (b) Initial linear wiping motions were generated, slanted due to estimation errors [$^{\circ}$]. The gripper then contacted the target object, executing three reciprocating wiping motions while regulating the contact force. The step numbers at the top left of the images correspond to those in~\figref{process}.}}
    \figlab{overview}
\end{figure*}
%%%%%%%%%%%%%%%%%%%%%%%%%%%%%%%%%%%%%%%%%%%%%%%%%%
\figref{system}~(a) displays the robot gripper with an Active Vapor sensing system. A two-fingered gripper (SAKE Robotics, EZGripper) serves as the end-effector, allowing the robot to grasp a wiping tool like a washcloth or sponge. A small ultrasonic humidifier atomizes the water, and the vibrator's ultrasonic waves cause the water to be ejected from the donut-shaped tip. An RGB camera, mounted at the bottom of the gripper, captures the target plane misted by the vapor. The camera operates at HD resolution (1280 x 720) and 30 fps. All devices are controlled by a single computer.

The camera captures the target plane, and the misted area is isolated using \textit{GrabCut}~\cite{GrabCut}. In the wiping experiments, the misted area's shape, identified as a cross after GrabCut processing, is used to estimate the normal of the misted surface.

\subsection{Contact-Regulated Robotic Wiper}
Stable contact with the target object's surface is necessary to effectively wipe off dirt. Instead of directly controlling the contact force, we regulate it within an empirically determined range with minimum and maximum normal force thresholds of 3.0 and 8.0 [N]. In our regulation process, the normal force $F_n$ is calculated as $-f_z$, representing the contact force in the z-axis direction.  

Experiments utilized a contact-force-regulated robotic wiping system. The contact force $f_z$ is the forward force when the robot approaches the target, derived from the torque sensors in the arm's six joints. Force values were obtained using \textit{iiwa\_stack}\footnote{https://github.com/IFL-CAMP/iiwa\_stack}, the ROS package for the KUKA LBR iiwa 14 R820 robot arm used in our experiments.

\section{Experiments}
To assess the effectiveness of the Active Vapor method, we conducted plane normal estimation for both a mirror and a transparent glass window. The experimental setup, depicted in \figref{system}~(b), involved attaching the sensor system to a seven-degree-of-freedom robot arm. An automatic rotating stage adjusted the target plane's azimuth angle. We used a 3mm thick mirror and glass, as the system cannot vaporize surfaces thicker than 3mm. Only the azimuth angle was estimated, varying by $-20^{\circ}$, $0^{\circ}$, and $20^{\circ}$, with three trials per pattern per object. Mist generation was challenging beyond 100mm from the target, so $r$ was set to 100mm. From this distance, the robot moved linearly with three vertical and horizontal reciprocations.

We performed ten trials to measure the time until the mist disappeared. Room temperatures were 23$^\circ$C, 26$^\circ$C, and 22$^\circ$C, respectively. The misted area vanished within approximately 6 seconds. Based on these results, a circular arc trajectory for the gripper was designed to complete within $T_{completion}$ = 6s. The robot gripper speed was set to 50mm/s to ensure the spraying and image capture were completed before the mist dissipated, exceeding the minimum calculated speed of 17mm/s.

\subsection{Plane Normal Estimation Accuracy}
We trained the GrabCut extraction model using 100 images from the same experimental environment. The RMSE of the azimuth estimation from $-20^{\circ}$ to $20^{\circ}$ was $4.2^{\circ}$ for the mirror and $5.8^{\circ}$ for the transparent glass window. These results indicate that the proposed Active Vapor-based sensing system, when attached to the robot, can accurately estimate the plane normal.

The smaller estimation error on the mirror surface is likely due to the laboratory setting, where few characteristic objects were in the background of transparent objects, making it challenging to extract the misted area. Conversely, the mirror surface reflected the robot gripper or arm, providing multiple objects with visual characteristics, allowing GrabCut to learn the misted area as the foreground based on refraction.
%%%%%%%%%%%%%%%%%%%%%%%%%%%%%%%%%%%%%%%%%%%%%%%%%%
\begin{figure}[t]
    \centering
    \begin{minipage}[t]{0.49\linewidth}
      \centering
      \includegraphics[width=\linewidth]{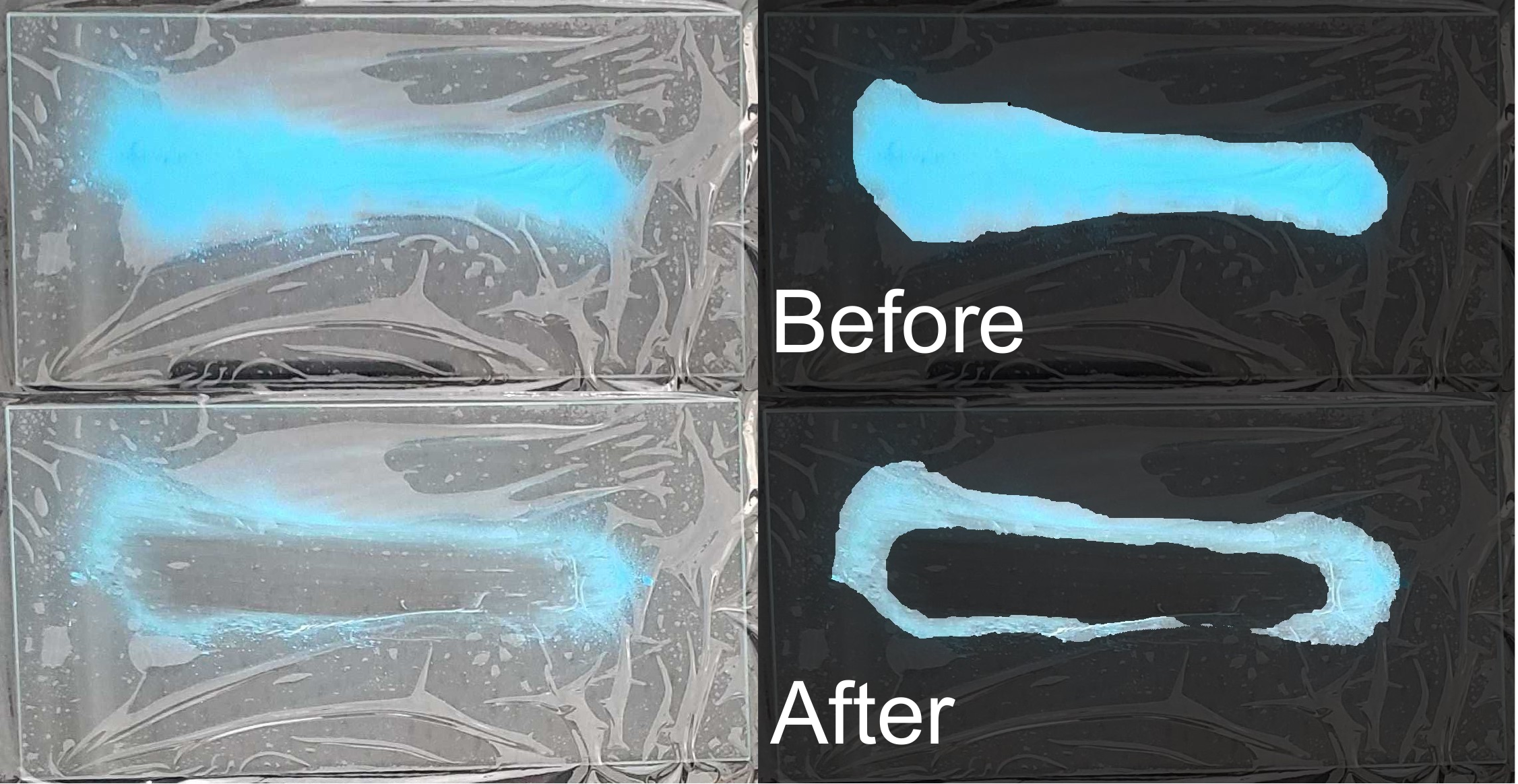}
      \subcaption{\small{Wiping result by a human}}
    \end{minipage}
    \begin{minipage}[t]{0.49\linewidth}
      \centering
      \includegraphics[width=\linewidth]{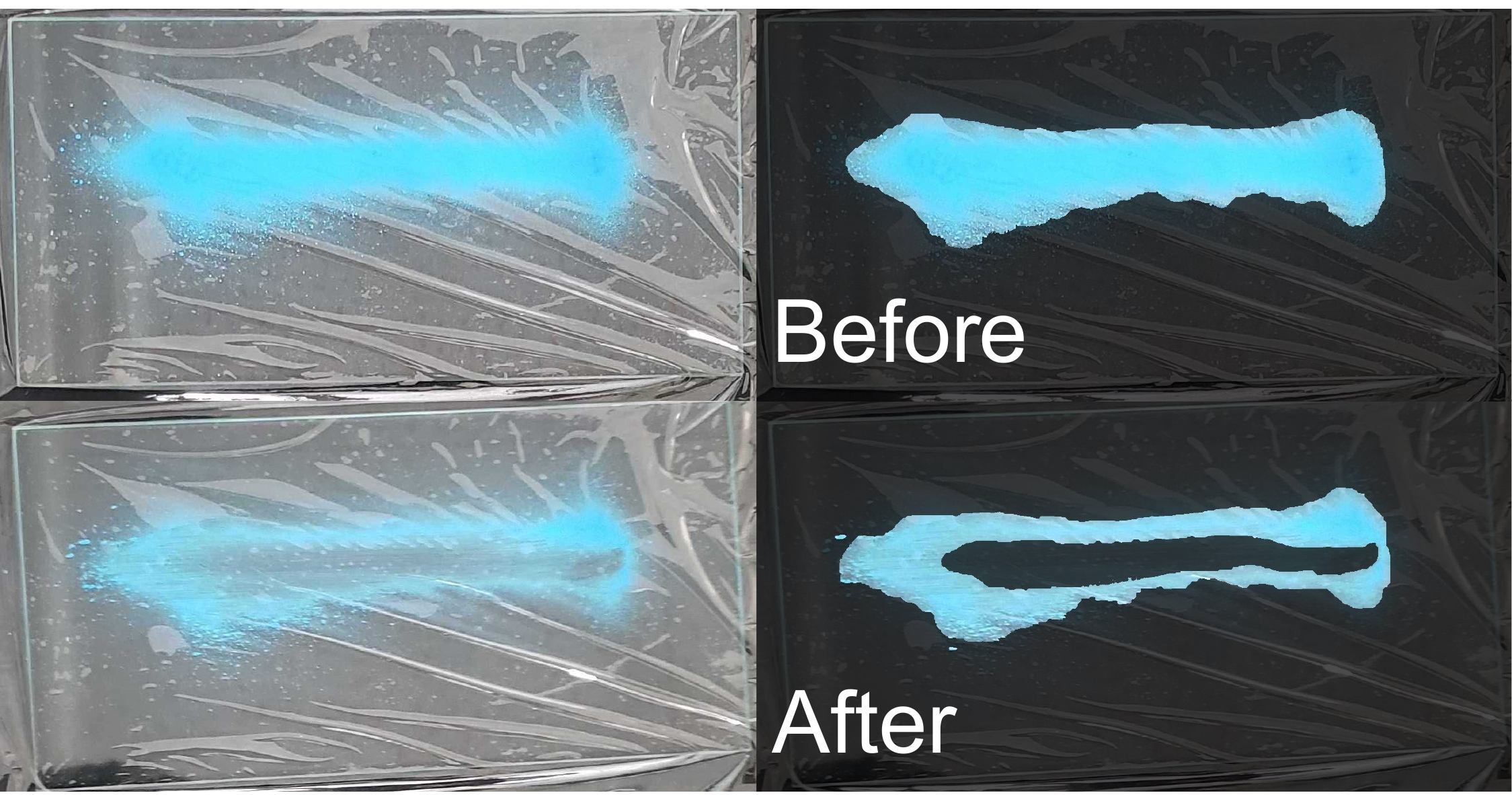}
      \subcaption{\small{Wiping result by a robot}}
    \end{minipage}
    \caption{\small{Ink-stained windows before and after wiping. Both (a) and (b) show the windows before a wiping motion (top left),  the result of segmenting the blue from the background (top right), the window after the wiping (bottom left), and the result (bottom right).}}
    \figlab{result_wiped}
\end{figure}
%%%%%%%%%%%%%%%%%%%%%%%%%%%%%%%%%%%%%%%%%%%%%%%%%%

\subsection{Evaluating Window Wiping}
\subsubsection{Wiping Accuracy}
The wiping performance was evaluated by manually painting water-based blue ink on a predefined window area of approximately 150mm by 20mm. The robot performed three round trips along a straight line (\figref{overview}~(b)). The remaining ink area was measured to determine performance, defined as $A_{unwiped}$ [$\mathrm{mm^2}$]. The relationship between window-wiping quality and azimuth angle estimation error was verified. The unwiped area $A_{unwiped}$ [$\mathrm{mm^2}$] is calculated as:

\begin{equation}
    A_{unwiped} = A_{initial} \times \frac{N_{final}}{N_{initial}},
\end{equation}
where $A_{initial}$ [$\mathrm{mm^2}$] is the initial ink-stained area (5000 $\mathrm{mm^2}$), and $N_{initial}$ and $N_{final}$ are the pixel counts of the ink-stained area before and after wiping, respectively.

The end effector moved along a straight line (\figref{overview}~(b)). The ink-stained area was extracted using GrabCut. We evaluated human wiping and robot wiping with a $5.8^{\circ}$ error, comparing the wiped area $\alpha$ [$\%$] for both cases.
\begin{equation}
    \alpha = \frac{(A_{initial} - A_{unwiped})}{A_{initial}} \times 100.
\end{equation}

\figref{result_wiped} presents the experimental results. With an azimuth angle estimation error of $5.8^{\circ}$ for the transparent glass window, the mean $\alpha$ across three trials was 45.2$\%$. The regulated contact force of the robot gripper enhanced the accuracy of the robot's wiping motions on the ink-stained surface, improving $\alpha$ to 52.2$\%$. When performed by a human, the mean $\alpha$ was 65.1$\%$. Thus, using an estimation error of $5.8^{\circ}$, approximately 52.2$\%$ of the ink can be removed.

\subsubsection{Wiping Efficiency}
We evaluated the efficiency of the proposed wiping system by measuring the reduction in time for wiping operations. We tested normal estimation with various spraying and image-capturing motions in a shorter duration. Our experiments indicated that a one-way spraying motion under 2 seconds typically fails, with failure defined as a misted area of nearly zero, confirmed by visual inspection. While results vary with camera frame rate, image-capturing motions under 3 seconds generally fail, when using a camera with a 30 fps frame rate.
%%%%%%%%%%%%%%%%%%%%%%%%%%%%%%%%%%%%%%%%%%%%%%%%%%
\begin{figure}[t]
  \centering
  \begin{minipage}[t]{0.32\linewidth}
    \centering
    \includegraphics[width=\linewidth]{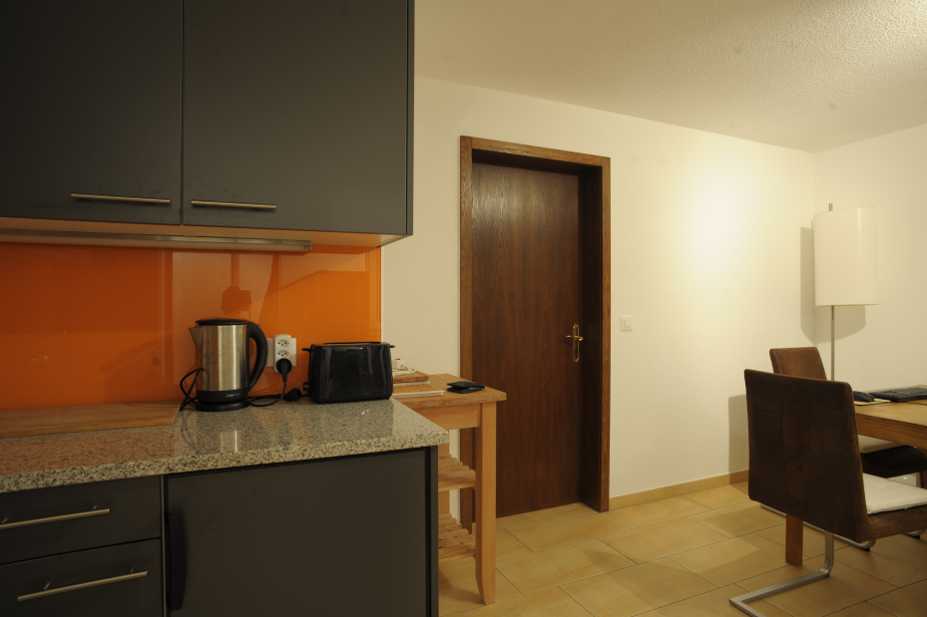}
    \subcaption{\small{Indoor scene}}
  \end{minipage}
  \begin{minipage}[t]{0.32\linewidth}
    \centering
    \includegraphics[width=\linewidth]{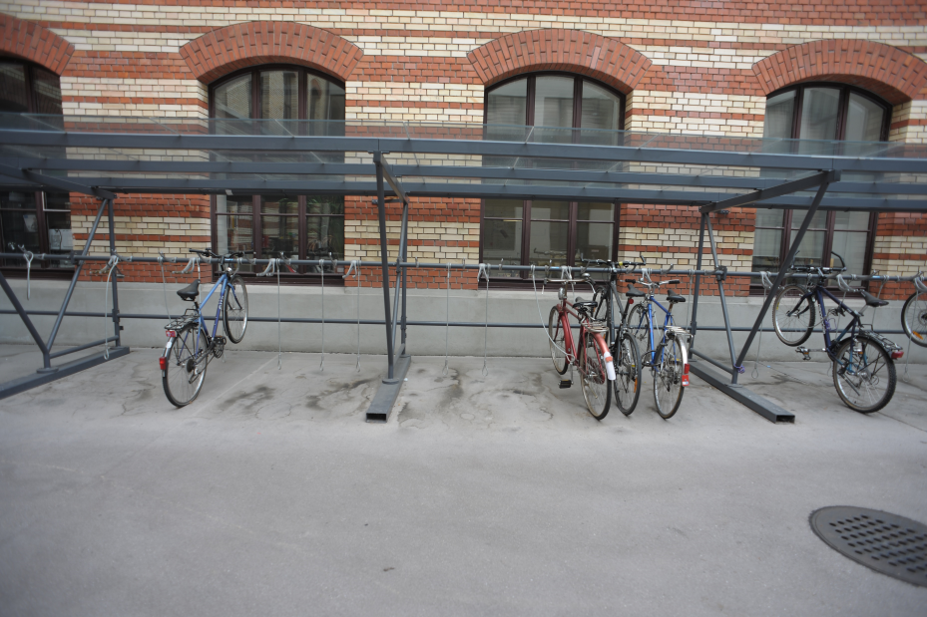}
    \subcaption{\small{Outdoor scene}}
  \end{minipage}
  \begin{minipage}[t]{0.32\linewidth}
    \centering
    \includegraphics[width=\linewidth]{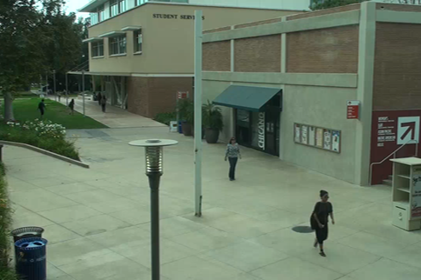}
    \subcaption{\small{Video frames}}
  \end{minipage}
  \caption{\small{Background scene examples. We used them to evaluate the robustness of the proposed misted-area extraction.}}
  \figlab{background}
\end{figure}
%%%%%%%%%%%%%%%%%%%%%%%%%%%%%%%%%%%%%%%%%%%%%%%%%%

\subsubsection{Wiping Effectiveness}
To validate the proposed wiping system's effectiveness, its accuracy was compared. Since accurately extracting misted areas based on the reflected scene and background was challenging, we tested the normal estimation's robustness on various backgrounds with a transparent window.

We compared the extraction accuracy of misted areas across different backgrounds for a transparent glass window, using indoor and outdoor scenes. Indoor (room) and outdoor (garden and sky) images were sourced from the \textit{ETH3D Stereo Benchmarks}~\cite{Schoeps2017cvpr}, and outdoor videos from the \textit{VIRAT Video Dataset}~\cite{Oh2011cvpr}. We used three images of different indoor scenes, three outdoor scenes, and three video frames of outdoor scenes. \figref{background} depicts the image examples used in the experiment.

\figref{result_bg} compares misted area extraction across different backgrounds. F-scores were calculated to evaluate segmentation accuracy using manually annotated ground truth data with \textit{labelme}\footnote{https://github.com/wkentaro/labelme}.
Five images per angle were annotated, and F-scores were calculated using the ground truth and our method's extracted misted areas.
The normal estimation error was $5.0^{\circ}$, slightly lower but similar to the normal estimation on a transparent glass plane with a textureless background. However, misted area extraction was more robust to viewpoints than the textureless background.

%%%%%%%%%%%%%%%%%%%%%%%%%%%%%%%%%%%%%%%%%%%%%%%%%%
\begin{figure}[t]
    \centering
    \begin{minipage}[t]{0.49\linewidth}
      \centering
      \includegraphics[width=\linewidth]{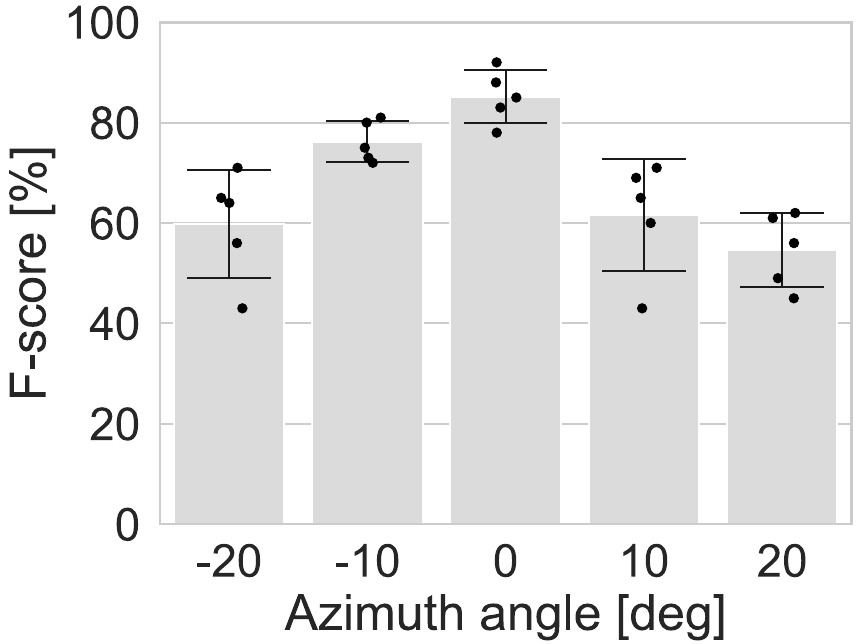}
      \subcaption{\small{Textureless background}}
    \end{minipage}
    \begin{minipage}[t]{0.49\linewidth}
      \centering
      \includegraphics[width=\linewidth]{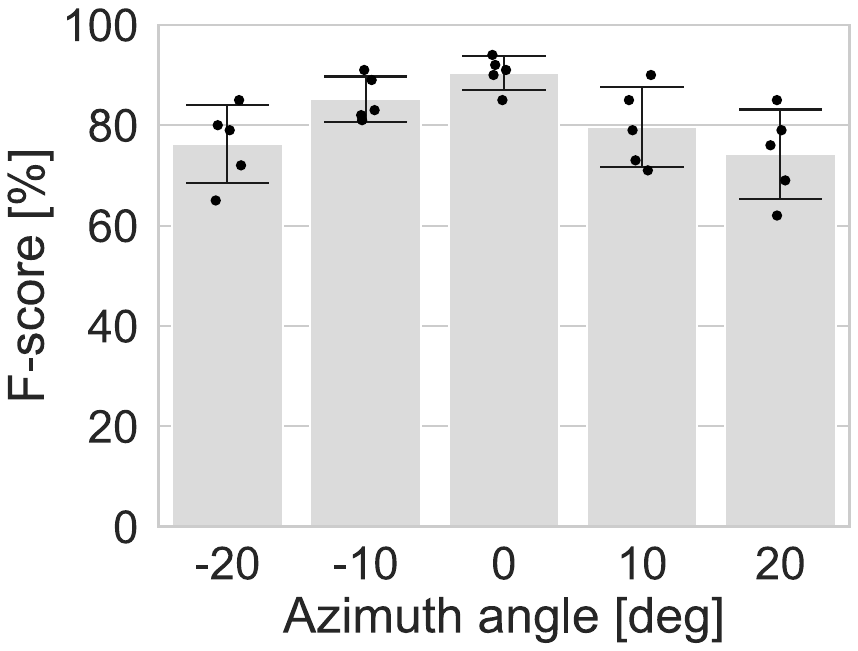}
      \subcaption{\small{With displayed real scenes}}
    \end{minipage}
    \caption{\small{Comparison of misted-area extraction from different backgrounds. Both (a) and (b) show the extraction results (F-scores). We annotated five images for each angle as the ground truth and calculated the F-scores using the extracted misted areas of the ground truth data and the extraction results.}}
    \figlab{result_bg}
\end{figure}
%%%%%%%%%%%%%%%%%%%%%%%%%%%%%%%%%%%%%%%%%%%%%%%%%%

\section{Discussion}
\subsubsection{Other Types of Vapor}
To enhance normal estimation accuracy, we considered painted water vapor for clearer identification of painted areas. We also explored using detergent, as it can be cleaned if it becomes dirty. However, spraying visible colored vapor on surfaces is challenging, and oil-based paint stains surfaces. Window-cleaning liquids are hazardous. Due to safety concerns, detergents were not used in this study.

\subsubsection{Controlling Contact}
Stable contact between the robot gripper and the target surface is crucial for accurate and efficient wiping motions. Leidner~\etal~\cite{Leidner2016} automated robot wiping using whole-body motion control, and Luo~\etal~\cite{Lu2015} allowed users to adjust the end-effector's force setpoint. Although force control is beyond this study's scope, these studies inform flexible wiping motion. Precise contact force control can increase motion time. Future work will focus on control methods and re-estimation of the plane normal to reduce motion time.

\subsubsection{Temperature}
The saturated water vapor content determines an adequate misted area for normal estimation. We tested drying times at room temperatures of 20°C, 25°C, and 30°C with 65\% humidity. The average drying times were 92s, 74s, and 68s, respectively, with minor, statistically insignificant differences.

\subsubsection{Improved Normal Estimation}
An algorithm to analyze reflections and background for optimal spraying is advisable due to the difficulty of extracting misted areas. Discussing a method for generating the minimum trajectory before the misted areas dry is also warranted. This study used a cross-shaped misted area, which is time-consuming; future work should optimize motions to save time. While this study focuses on plane normal estimation, normal estimation for curved surfaces using light-field features~\cite{LightField} is a promising future direction.
%%%%%%%%%%%%%%%%%%%%%%%%%%%%%%%%%%%%%%%%%%%%%%%%%%
\begin{figure}[t]
  \centering
  \begin{minipage}[t]{0.49\linewidth}
    \centering
    \includegraphics[width=\linewidth]{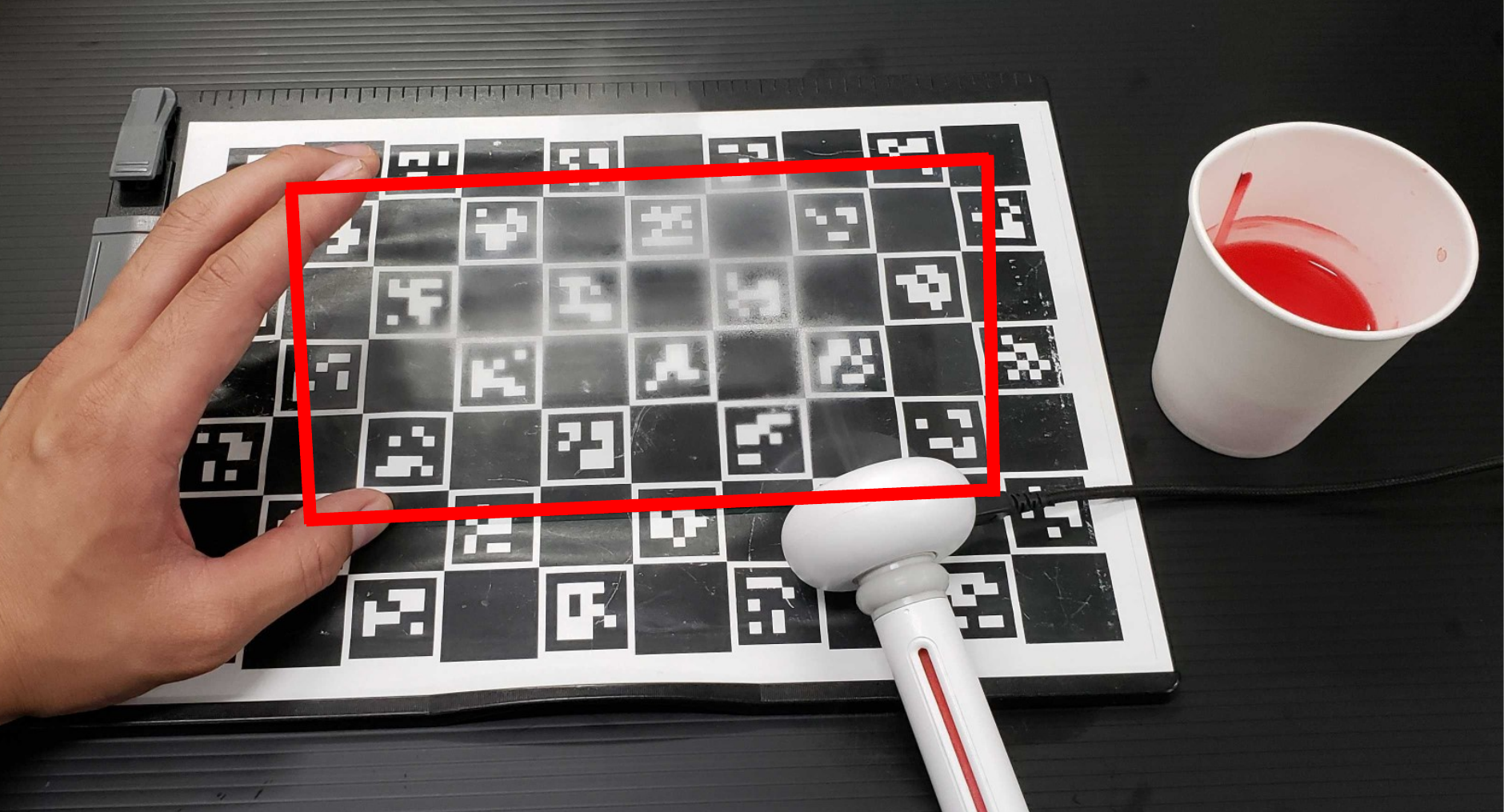}
    \subcaption{\small{Misted glass with red water}}
  \end{minipage}
  \begin{minipage}[t]{0.49\linewidth}
    \centering
    \includegraphics[width=\linewidth]{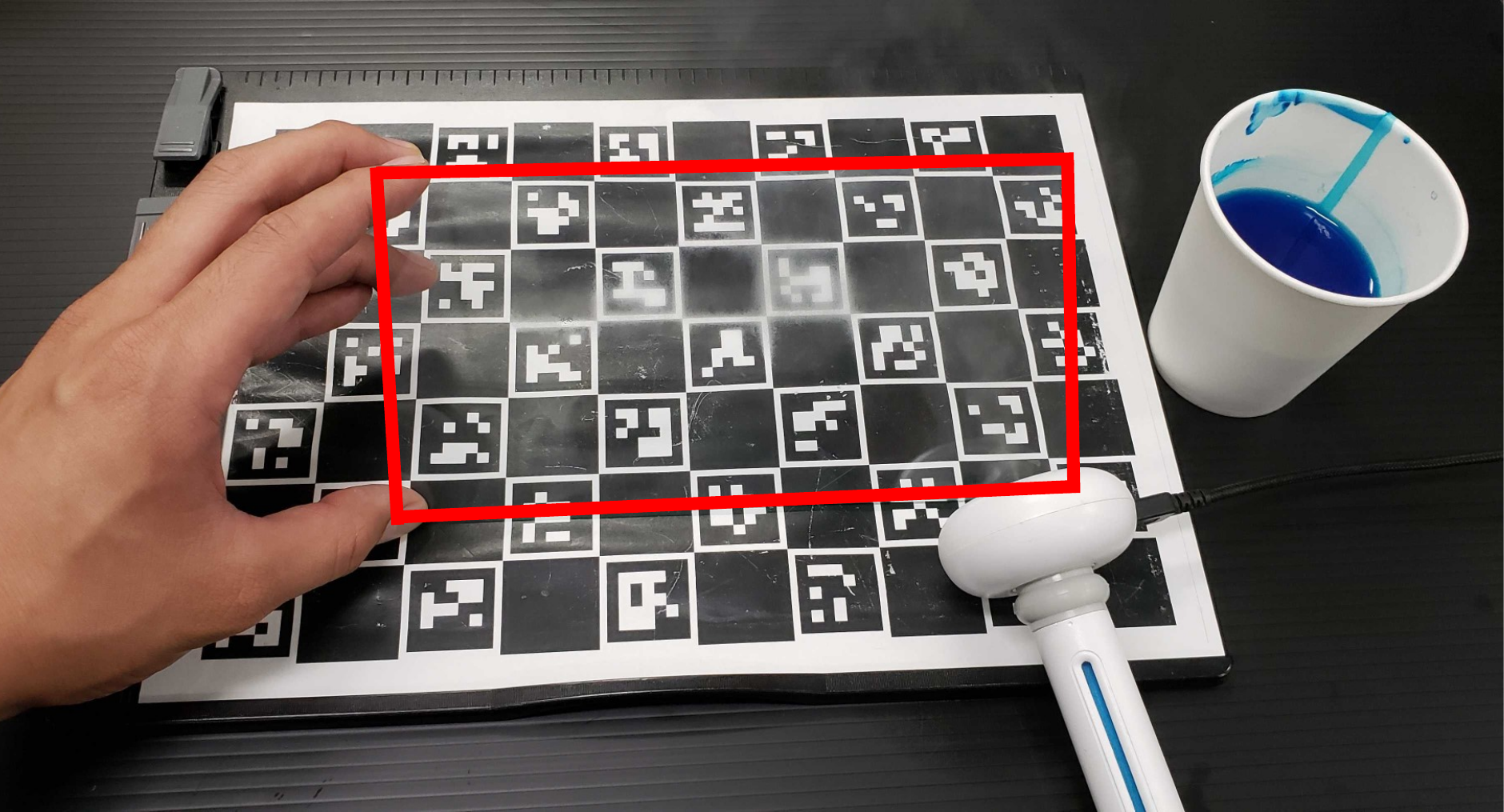}
    \subcaption{\small{Misted glass with blue water}}
  \end{minipage}
  \caption{\small{Surfaces sprayed of colored water vapor. We colored the water and saturated the sponge inside the humidifier with colored water. The humidifier then sprayed water vapor.}}
  \figlab{painted}
\end{figure}
%%%%%%%%%%%%%%%%%%%%%%%%%%%%%%%%%%%%%%%%%%%%%%%%%%

\section{Conclusions}
This study proposes a method to estimate the normals of mirrors and transparent objects that are challenging to recognize with a camera. A sensor system on a robot arm wipes the target object, and water vapor is sprayed to create a cross-shaped misted area, which is extracted using GrabCut to estimate the plane normal. Experiments showed an average error of $4.2^{\circ}$ for mirrors and $5.8^{\circ}$ for transparent glass windows.
 
Future work will enhance the estimation system, test various home objects, and develop a contact-controlled robotic window-cleaning system robust to normal estimation errors.

\bibliographystyle{IEEEtran}
\footnotesize
\bibliography{reference}

\end{document}